\pdfoutput=1
\documentclass[11pt]{article}

\usepackage[final]{acl}

\usepackage{times}
\usepackage{latexsym}
\usepackage{booktabs}
\usepackage{enumitem}
\usepackage{algorithm}
\usepackage{algpseudocode}
\usepackage{tcolorbox}
\usepackage{titlesec}
\tcbuselibrary{breakable}
\tcbuselibrary{listings}
\usepackage{caption}  
\usepackage{graphicx}
\usepackage{subcaption}
\usepackage{multirow}
\usepackage[T1]{fontenc}
\usepackage[utf8]{inputenc}

\usepackage{microtype}
\usepackage{fontawesome5}
\usepackage{inconsolata}

\newcommand{\diagrams}{\textsc{Diagrams}}

\usepackage{placeins}

\tcbset{
  verticalbox/.style={
    enhanced,
    colback=#1!6,
    colframe=#1!60!black,
    boxrule=0.6pt,
    arc=2mm,
    left=1.5mm,right=1.5mm,top=1mm,bottom=1mm,
    before skip=0.6\baselineskip,
    after skip=0.6\baselineskip,
  }
}

\title{\diagrams : A Review Framework for Reasoning-Level Attribution in Diagram QA}

\newcommand{\demolink}{\href{https://coral-lab-asu.github.io/diagrams/}{\faIcon{play-circle}\hspace{2pt}Project Page}}
\newcommand{\videolink}{\href{https://youtu.be/oli_ekuLYjo}{\faIcon{video}\hspace{2pt}Video}}
\newcommand{\codelink}{\href{https://github.com/CoRAL-ASU/diagrams}{\faIcon{github}\hspace{2pt}Code}}

\newcommand{\logosup}[1]{\raisebox{0.6ex}{\hbox{\scriptsize #1}}}
\newcommand{\asulogo}{\includegraphics[height=8pt]{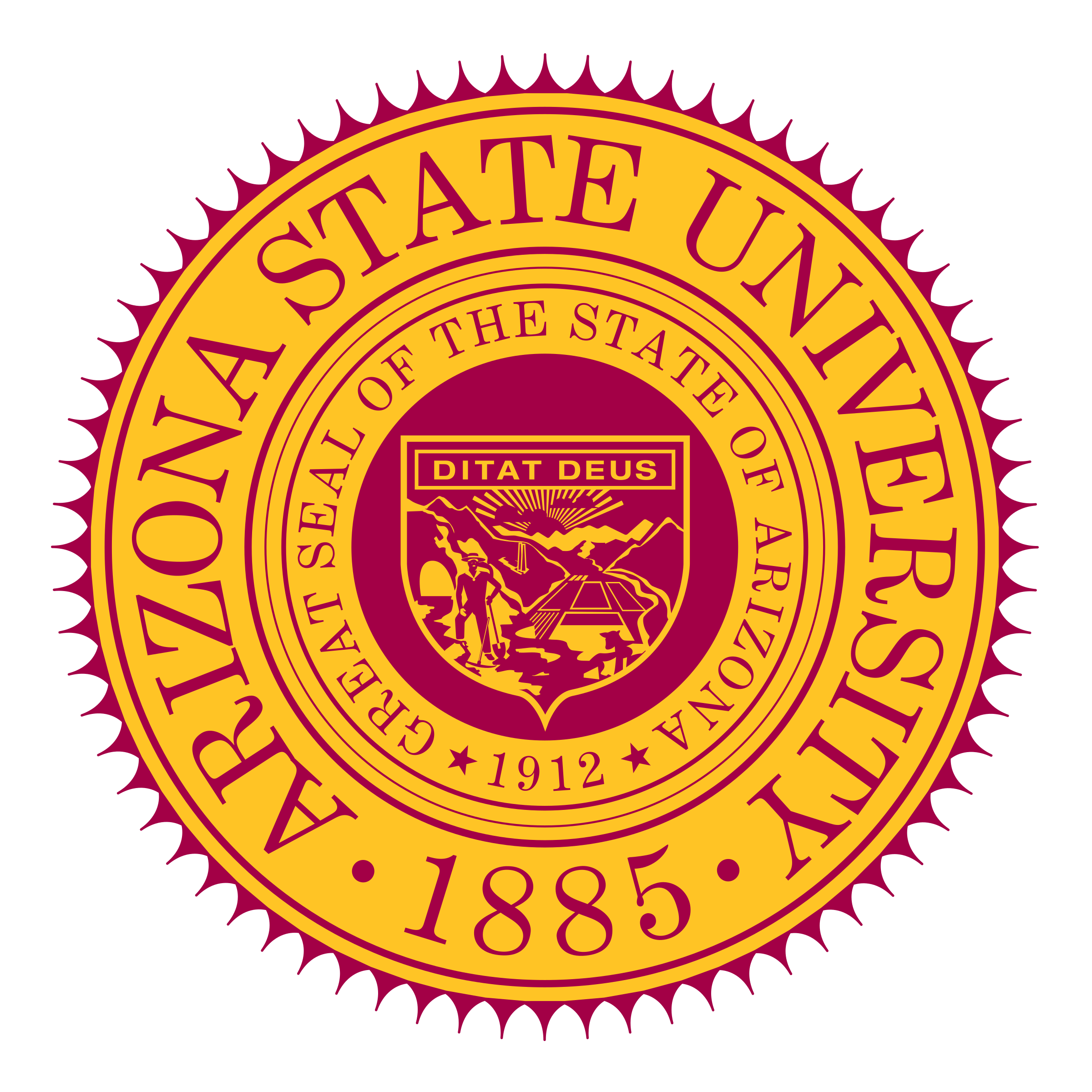}}
\newcommand{\iitdmlogo}{\includegraphics[height=8pt]{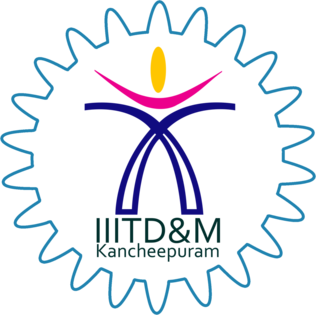}}
\newcommand{\adobelogo}{\includegraphics[height=8pt]{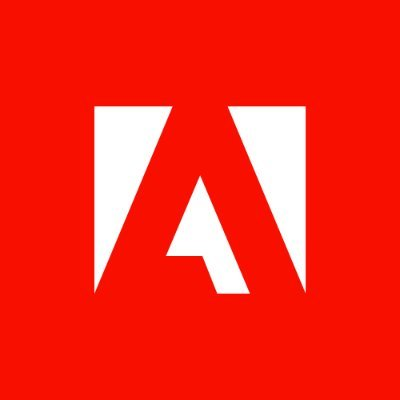}}
\newcommand{\umclogo}{\includegraphics[height=8pt]{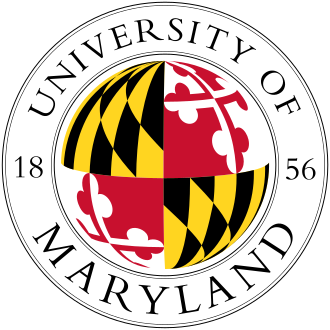}}

\newcommand{\asu}{\logosup{\asulogo}}
\newcommand{\iiitdm}{\logosup{\iitdmlogo}}
\newcommand{\adobe}{\logosup{\adobelogo}}
\newcommand{\umc}{\logosup{\umclogo}}

\author{
\textbf{Anirudh Iyengar Kaniyar Narayana Iyengar}\asu\textsuperscript{*},
\textbf{Tampu Ravi Kumar}\asu\textsuperscript{*},\\
\textbf{Manan Suri}\umc,
\textbf{Raviteja Bommireddy}\iiitdm,
\textbf{Dinesh Manocha}\umc,\\
\textbf{Puneet Mathur}\adobe\textsuperscript{*},
\textbf{Vivek Gupta}\asu\textsuperscript{*}
\\[2pt]
\asu~Arizona State University 
\adobe~Adobe Research 
\iiitdm~IIITDM 
\umc~University of Maryland
\\[2pt]
\demolink \quad
\videolink \quad
\codelink
\\[2pt]
\texttt{\footnotesize akaniyar@asu.edu, traviku2@asu.edu,
manans@umd.edu, cs23b2011@iiitdm.ac.in}\\[-4pt]
\texttt{\footnotesize
dmanocha@umd.edu,
puneetm@adobe.com, vgupt140@asu.edu
}
}

\begin{document}
\maketitle

\begin{abstract}
Diagram question answering (Diagram QA) requires \emph{reasoning-level attribution} that links each question–answer pair to all visual regions needed to derive the answer, rather than only the region containing the final response. Creating such structured evidence across diagrams, charts, maps, circuits, and infographics is time-consuming, and existing annotation tools tightly couple their interfaces to dataset-specific formats. We present \textbf{\diagrams}, a lightweight, schema-driven review framework that decouples interface logic from dataset-specific JSON structures through an internal meta-schema and dataset adapters. Given an image and QA pair with optional candidate regions, the system performs QA-conditioned evidence selection and proposes the regions required for reasoning. When QA pairs or candidate regions are missing, it generates them and supports human verification and refinement.  Across six Diagram QA datasets, model-suggested evidence achieves 85.39\% precision and 75.30\% recall against reviewer-final selections (micro-averaged). These results indicate that the review-first framework reduces manual region creation while maintaining high agreement with final reasoning-level attributions. We release a public demo and installable package to support dataset auditing, grounded supervision creation, and grounded evaluation.
\end{abstract}

\begin{figure*}[t]
\centering
\includegraphics[width=0.9\linewidth]{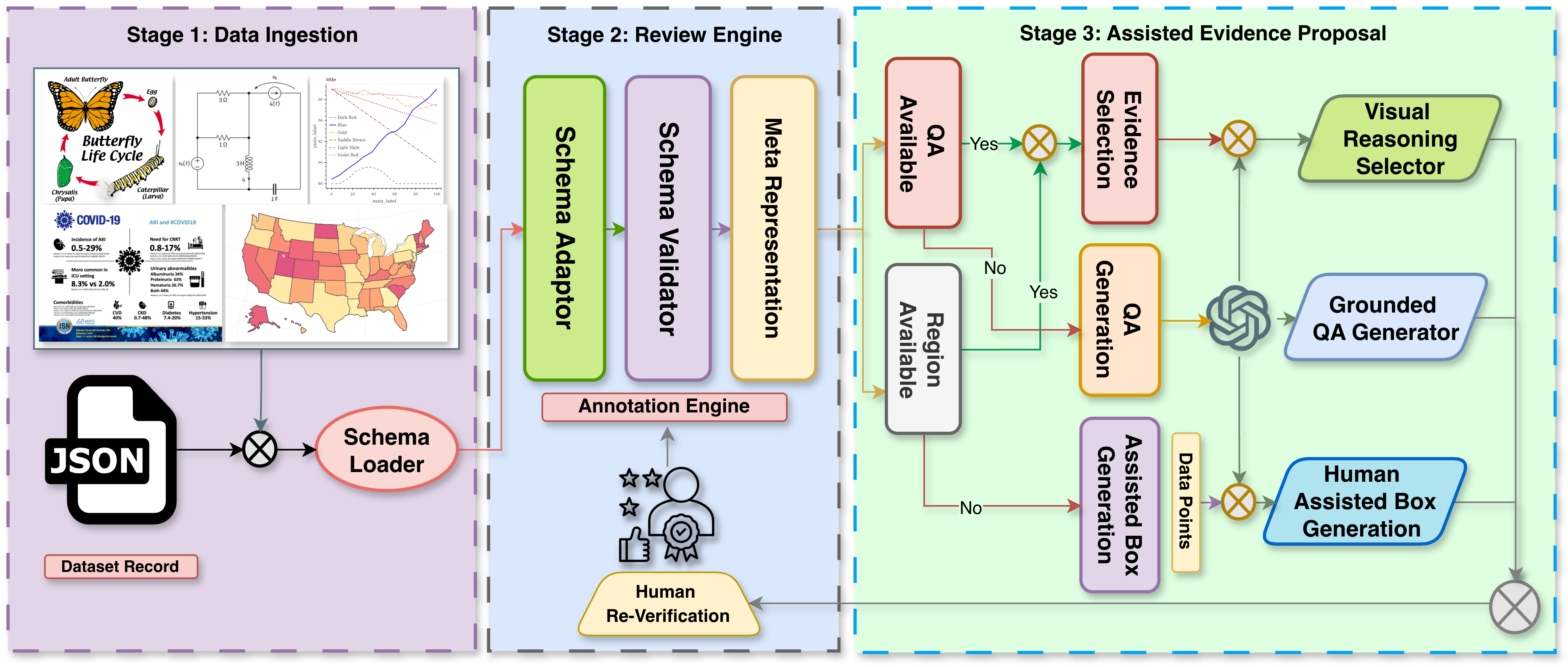}
\caption{Overview of the \textbf{\diagrams{}} architecture. The framework operates in three stages: (1) Data Ingestion, which converts dataset-specific records into a unified meta-schema; (2) Review Engine, which renders normalized QA items and candidate regions for human verification; and (3) Assisted Evidence Proposal, where a multimodal backend performs QA-conditioned evidence selection and optional QA or region generation. }

\label{fig:architecture}
\end{figure*}

\section{Introduction}
Diagram question answering (Diagram QA) spans diverse visual domains, including scientific diagrams~\cite{kembhavi2016diagram}, charts~\cite{masry2022chartqa}, circuit schematics~\cite{mehta2024circuitvqa}, infographics~\cite{mathew2022infographicvqa}, and map-based benchmarks~\cite{srivastava2025mapiq,mukhopadhyay2025mapwise}. Unlike pure text-based reasoning, Diagram QA requires models to interpret structured visual elements and relate them to linguistic queries. In many cases, answering a question does not depend on a single visual region, but on a chain of interconnected elements. For example, when a question asks about the neighbors of a highlighted state on a map, the reasoning process requires identifying the target state and examining all adjacent regions. The final answer emerges from multiple visual dependencies rather than from a single answer-containing box.

Despite this, most existing annotation efforts focus on answer-level grounding, where annotators mark only the region that directly contains the final response. Such annotations fail to capture the full reasoning path that supports the answer. Reasoning-level attribution links each question–answer pair to all visual regions required to derive the answer. This richer supervision enables grounded training, faithfulness evaluation, and systematic error analysis of vision-language models (VLMs). Without structured attribution, models may produce correct answers while relying on spurious correlations or hallucinated associations. 
Recent analyses of chart reasoning benchmarks demonstrate that models often struggle with distributed and compositional visual evidence, reinforcing the need for structured, region-level attribution~\cite{Iyengar2025InterChartBV}. Annotators find reasoning-level attribution expensive. Annotators must analyze the diagram, identify all relevant regions, and create or refine precise bounding boxes. The burden increases further because Diagram QA datasets vary widely in JSON structure, coordinate conventions, region availability, and QA formatting. Many existing annotation interfaces tightly couple their design to a specific dataset format, making it difficult to reuse tools across heterogeneous resources. As a result, researchers often rebuild front-end logic for each new dataset, slowing progress and limiting reproducibility.


We introduce \textbf{\diagrams}, a review-first framework for reasoning-level attribution in Diagram QA. Rather than centering the interface on dataset-specific schemas, \diagrams{} converts each input record into an internal meta-schema representation that decouples annotation logic from dataset structure. A multimodal backend proposes QA-conditioned evidence regions, and the human reviewer verifies, edits, removes, or adds regions as needed. When candidate boxes or QA pairs are missing, the system proposes them and supports interactive refinement. \diagrams{} targets researchers developing Diagram QA benchmarks, practitioners building grounded supervision datasets, and vision-language researchers studying reasoning-level attribution.

We evaluate \diagrams{} across six Diagram QA datasets spanning charts, maps, diagrams, circuits, and infographics. Model-suggested evidence achieves high agreement with reviewer-final selections, demonstrating that a review-first pipeline substantially reduces manual effort while preserving reasoning-level fidelity.

\noindent We summarize our contributions as follows:
\begin{enumerate}
\setlength\itemsep{0em}
\item We present \textbf{\diagrams{}}, a review framework that formalizes \emph{reasoning-level} attribution for Diagram QA, explicitly distinguishing it from answer-only grounding.
\item We design a multimodal proposal pipeline that performs QA-conditioned evidence selection, missing-region generation, and optional QA generation, enabling efficient human-in-the-loop verification.
\item We provide empirical evaluation across six visual QA datasets, demonstrating strong proposal utility under a review-first workflow.
\end{enumerate}


\section{Related Work}
Existing image annotation platforms provide extensive support for drawing and editing visual regions such as bounding boxes, polygons, and masks. Systems including LabelMe~\cite{russell2008labelme}, VIA~\cite{dutta2019via}, CVAT~\cite{sekachev2019computer}, and Label Studio~\cite{tkachenko2020label} also incorporate model-assisted workflows to accelerate labeling. These platforms standardize region creation and editing but operate primarily at the object or instance level rather than supporting \emph{QA-conditioned} evidence attribution, and they do not address the structural heterogeneity inherent in Diagram QA datasets. Prior work has further reduced bounding-box effort through interaction techniques such as boundary snapping (Snapper)~\cite{williams2024snapper} and iterative human-in-the-loop box refinement~\cite{adhikari2021iterative}. In parallel, VQA research has explored answer grounding, attention supervision, and explanation alignment~\cite{das2017human, wang2020general, chen2022grounding, pantazopoulos2025towards, selvaraju2017grad, selvaraju2019taking}, often using region-level or attention-based signals to improve model interpretability. However, these approaches typically annotate answer-containing regions or saliency maps rather than full reasoning chains. \diagrams{} extends this line of work
by targeting \emph{reasoning-level} attribution and by providing a schema-driven review workflow that remains reusable across heterogeneous dataset formats.

\section{\diagrams{}: System Architecture and Workflow}
\diagrams{} operates in three coordinated stages: data ingestion and normalization, QA-conditioned evidence proposal, and structured human verification. The system first converts heterogeneous dataset records into a unified internal meta-schema representation. It then performs QA-conditioned evidence selection or generation using a multimodal backend. Finally, reviewers verify, refine, and export reasoning-level attributions through an interactive interface. Figure~\ref{fig:architecture} illustrates this end-to-end workflow.

\begin{figure}[t]
\centering
\includegraphics[width=\linewidth]{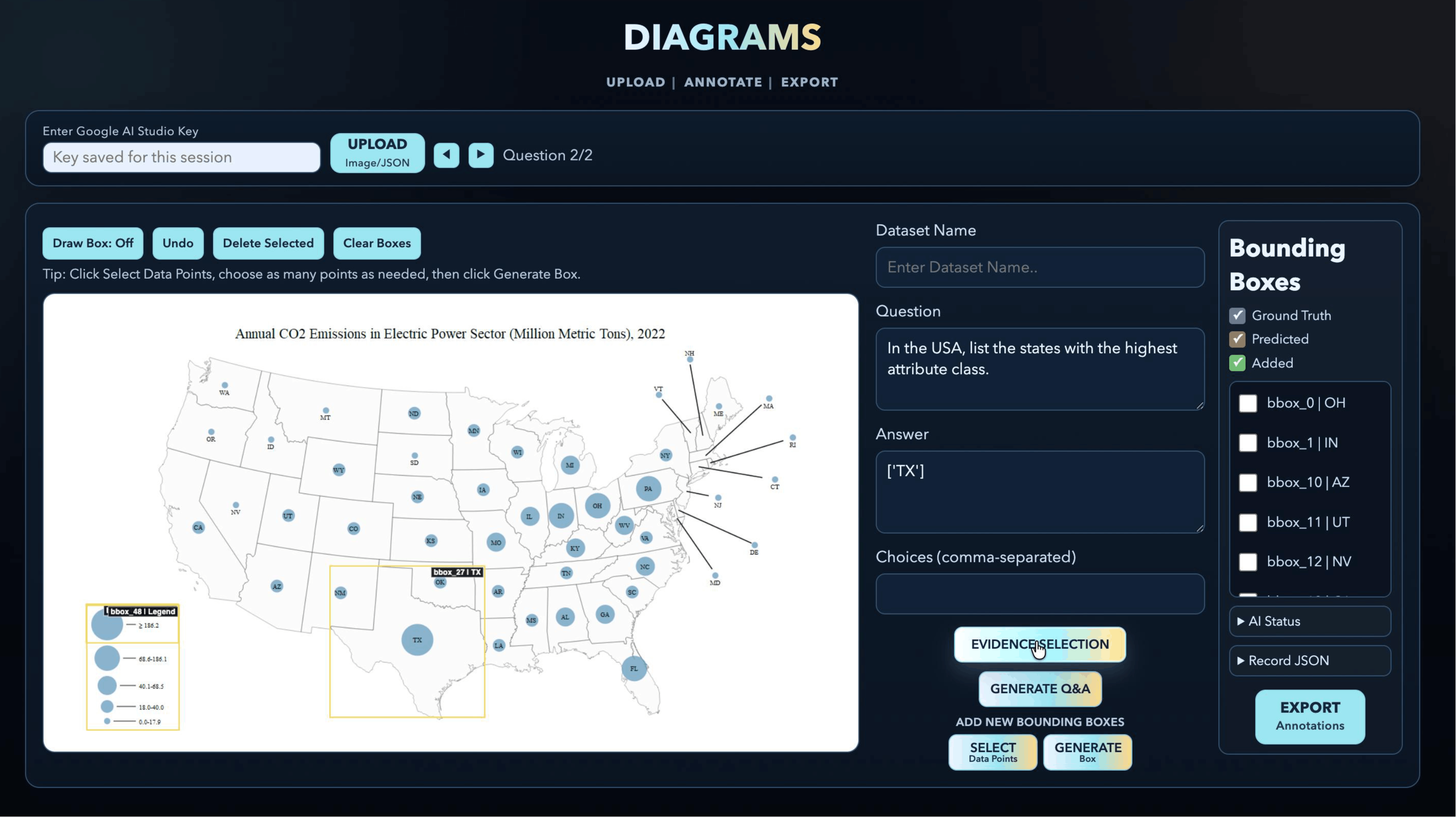} 
\caption{Selection-based attribution. The reviewer verifies QA-conditioned region proposals and inspects ground-truth, predicted, and added boxes.}
\label{fig:ui_map}
\end{figure}
\subsection{Architecture}

The framework consists of the following components:


\begin{enumerate}[leftmargin=*, itemsep=2pt, topsep=2pt, parsep=0pt, partopsep=0pt]

\item \textbf{Schema Loader}: The system ingests a JSON file and retrieves the corresponding images, questions/annotations.

\item \textbf{Schema Adapter}: The adapter maps dataset-specific fields into an internal meta-schema representation.

\item \textbf{Schema Normalizer}: The normalization layer standardizes heterogeneous bounding box formats (e.g., \texttt{[x,y,w,h]}, \texttt{left/top/width/height}).

\item \textbf{Internal Meta-Schema Representation}: The system stores images, QA pairs, candidate regions, and attribution mappings in a unified internal structure.
\item \textbf{Annotation Engine}: The interface renders images, QA items, and regions, and tracks edits through standard region operations.
\item \textbf{Evidence Proposal Module}: The multimodal backend performs QA-conditioned evidence selection or generation.
\item \textbf{Human Re-Verification}: The reviewer verifies, edits, removes, or adds regions and QA items.
\item \textbf{Exporter}: The system saves finalized annotations in structured JSON format and optionally exports annotated images.
\end{enumerate}
This modular design separates dataset-specific parsing from attribution logic and human interaction. By decoupling dataset-specific schema handling from attribution logic, the system remains reusable across heterogeneous Diagram QA datasets without requiring interface redesign.


\begin{figure}[t]
\centering
\includegraphics[width=0.9\linewidth]{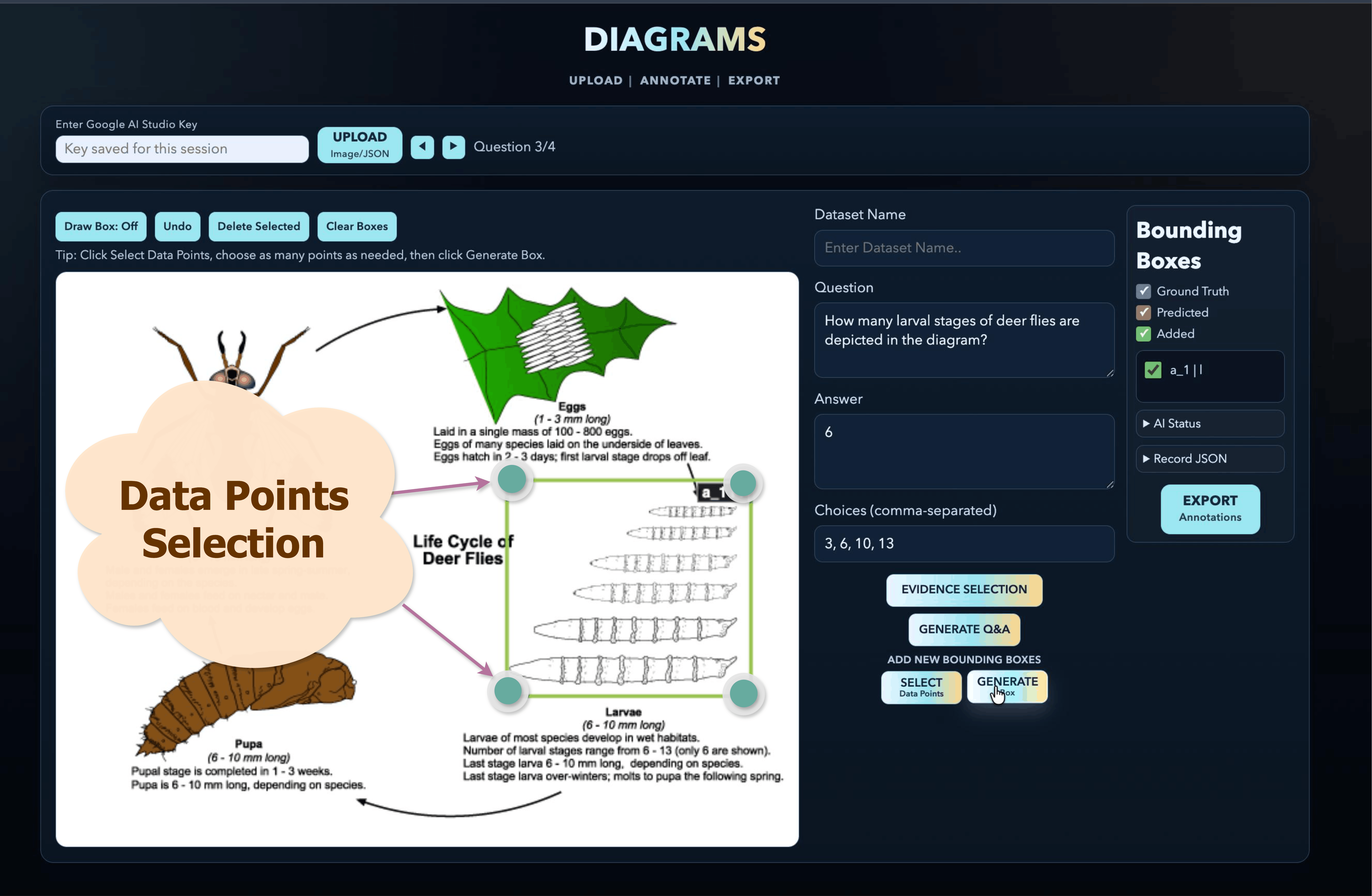}
\caption{Proposal-based attribution with region generation. The reviewer draws or refines evidence boxes when candidate regions are unavailable.}
\label{fig:ui_pie}
\end{figure}
\subsection{Review-First Workflow}

\begin{figure}[t]
\centering
\includegraphics[width=\linewidth]{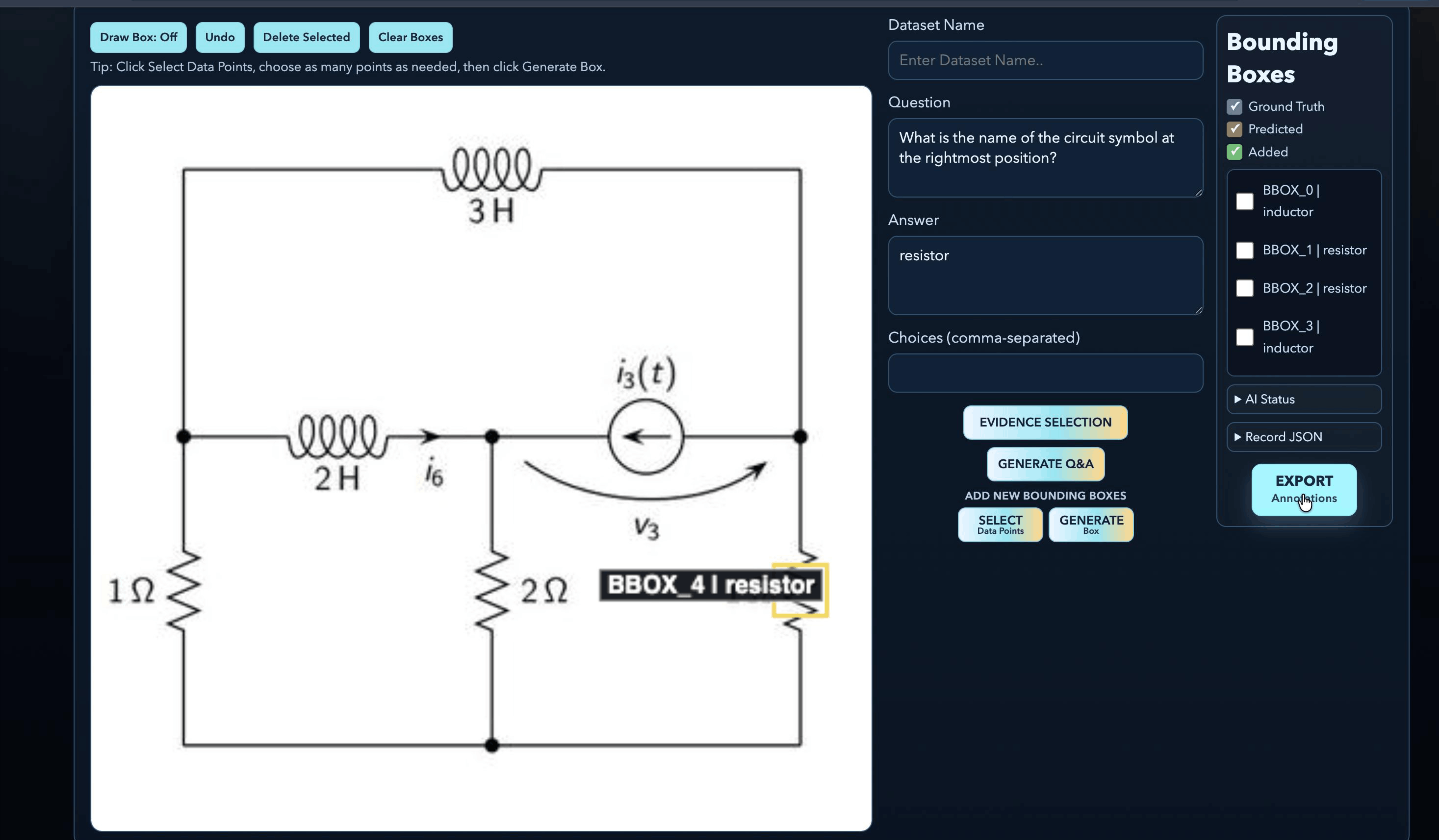}
\caption{Human verification and export. The reviewer confirms, edits, or removes proposed regions before finalizing the annotation.}
\label{fig:ui_bar}
\end{figure}

\diagrams{} implements a review-first workflow that prioritizes structured verification over manual region creation. After loading and normalizing a record, the system invokes the evidence proposal module to generate QA-conditioned region suggestions. The reviewer inspects these proposals directly in the interface and verifies, edits, or removes them instead of drawing all regions from scratch. This interaction model shifts the annotation process from box creation to guided verification. We illustrate both selection-based and proposal-based workflows in Figures~\ref{fig:ui_map}-\ref{fig:ui_bar}.

Depending on data availability, the system operates in two settings:

\textbf{Selection-based attribution:} When the dataset provides QA pairs and candidate regions, the system performs QA-conditioned evidence selection over the candidate region pool and initializes the selected subset in the interface.

\textbf{Proposal-based attribution:} When QA pairs and/or candidate regions are missing, the system proposes them automatically and supports interactive human refinement before finalization.






\section{Evidence Proposal}

\subsection{Evidence Selection from Candidate Regions}

When both QA pairs and candidate regions are available, the backend performs QA-conditioned evidence selection over the candidate region pool. Given the question, answer, and candidate bounding-boxes, the multimodal model proposes the subset of regions required to support the reasoning chain as shown in Figure~\ref{fig:ui_map}. The selected region identifiers are initialized in the interface. The reviewer verifies, edits, or augments the proposed evidence before export.

\subsection{Region Generation When Candidates Are Absent}

When candidate regions are unavailable but QA pairs exist, the backend proposes new evidence regions directly on the image. The reviewer may accept, edit, or delete these proposals and may draw additional regions if required. Figure~\ref{fig:ui_pie} illustrates manual refinement when candidate regions are absent or incomplete.

\subsection{QA and Region Generation}

When QA pairs and candidate regions are missing, the system first proposes QA items and then generates corresponding evidence regions. The reviewer verifies and refines both components before finalizing attribution, as shown in Figure~\ref{fig:ui_bar}.








\begin{table}[t]
\small
\centering
\setlength{\tabcolsep}{4pt}
\begin{tabular}{lccc}
\toprule
\textbf{Dataset} & \textbf{Precision} & \textbf{Recall} & \textbf{F1} \\
\midrule
ChartQA           & 54.79 & 42.22 & 47.69 \\
MapWise           & 94.49 & 64.10 & 76.38 \\
AI2D              & \textbf{99.64} & 91.13 & 95.19 \\
Circuit-VQA       & 66.97 & 49.50 & 56.92 \\
InfographicsVQA   & 96.49 & 98.97 & \textbf{97.72} \\
MapIQ             & 91.13 & \textbf{99.74} & 95.24 \\
\midrule
\textbf{Overall}  & \textbf{85.39} & \textbf{75.30} & \textbf{80.03} \\
\bottomrule
\end{tabular}
\caption{Proposal utility after human verification (micro-averaged). Precision = TP/(TP+FP), Recall = TP/(TP+FN), F1 = harmonic mean.}
\label{tab:proposal_metrics}
\end{table}

\section{Annotation Engine}

The annotation engine renders images, QA items, and regions in an interactive interface. Reviewers can select, resize, delete, or draw regions, while the system tracks edits and maintains links between QA items and evidence regions. The interface exposes normalized metadata to preserve traceability between the internal meta-schema and the original dataset record. Figure~\ref{fig:ui_map} shows selection-based attribution, where candidate regions exist and the reviewer verifies proposed evidence.

\section{Human Re-Verification and Export}

During re-verification, reviewers confirm or modify proposed QA items and evidence regions. The system records whether each region was proposed, edited, removed, or newly created. Appendix~\ref{appendix:annotation_instructions} details the annotation guidelines and arbitration procedures. After verification, the exporter saves the finalized attribution in structured JSON format and optionally produces an annotated image. These outputs support downstream training, auditing, and evaluation workflows. Figure~\ref{fig:ui_bar} shows reviewers finalizing attribution before export.

\section{Results and Analysis}
\label{sec:results}


We organize our evaluation around four research questions. (RQ1) Does QA-conditioned evidence proposal reduce manual annotation effort while maintaining correctness? (RQ2) Does proposal utility generalize across heterogeneous Diagram QA datasets? (RQ3) What types of errors drive reviewer intervention? (RQ4) Is proposal verification reproducible across annotators? We answer these questions through large-scale proposal utility analysis across six datasets. Specifically, we use AI2D~\cite{kembhavi2016diagram}, ChartQA~\cite{masry2022chartqa}, Circuit-VQA~\cite{mehta2024circuitvqa}, InfographicVQA~\cite{mathew2022infographicvqa}, MapIQ~\cite{srivastava2025mapiq}, and MapWise~\cite{mukhopadhyay2025mapwise} and a focused reliability study on ChartQA and Circuit-VQA.

\begin{table}[t]
\small
\centering
\setlength{\tabcolsep}{6pt}
\begin{tabular}{lcc}
\toprule
\textbf{Dataset} & \textbf{Added GT} & \textbf{New Drawn} \\
\midrule
ChartQA           & 8308 & 215 \\
MapWise           & 2700 & 1162 \\
AI2D              & 1173 & 23 \\
Circuit-VQA       & 1405 & 1846 \\
InfographicsVQA   & 28 & 68 \\
MapIQ             & 36 & 0 \\
\midrule
\textbf{Overall}  & \textbf{13650} & \textbf{3314} \\
\bottomrule
\end{tabular}
\caption{Breakdown of reviewer-added evidence regions. ``Added GT'' denotes missed ground-truth boxes; ``New Drawn'' denotes boxes created during review.}
\label{tab:fn_breakdown}
\end{table}

\subsection{Proposal Utility Across Six Datasets}
We evaluate proposal utility by comparing model-suggested regions to reviewer-approved evidence using micro-averaged precision, recall, and $F_1$. Appendix~\ref{appendix:Evaluation} provides formal metric definitions and implementation details.

\paragraph{Do model proposals reduce manual annotation effort?}
As shown in Table~\ref{tab:proposal_metrics}, the backend proposed 60,553 boxes, of which reviewers retained 51,709 (85.39\% precision). Reviewers finalized 68,673 evidence boxes in total, and proposals achieved 75.30\% recall and 80.03 \(F_1\) (micro-averaged). These results show that the system pre-selects a large fraction of final evidence regions, converting large-scale manual box creation into efficient verification.

\paragraph{Does the system prioritize correctness over over-generation?}High precision (85.39\%) indicates that reviewers removed relatively few suggested boxes. The backend avoids aggressive over-proposal and instead surfaces regions that reviewers largely accept, demonstrating stable QA-conditioned selection.

\paragraph{Where does the system still miss evidence?} Reviewers added 16,964 missing boxes (Table~\ref{tab:fn_breakdown}), indicating that recall remains the primary area for improvement. Among these, 13,650 correspond to missed ground-truth regions and 3,314 to newly drawn boxes, suggesting that missed candidate coverage drives most recall gaps.

\begin{table}[t]
\small
\centering
\setlength{\tabcolsep}{6pt}
\begin{tabular}{lc}
\toprule
\textbf{Dataset} & \textbf{New Drawn (\%)} \\
\midrule
ChartQA           & 2.52 \\
MapWise           & 30.10 \\
AI2D              & 1.92 \\
Circuit-VQA       & 56.79 \\
InfographicsVQA   & \textbf{70.83} \\
MapIQ             & 0.00 \\
\midrule
\textbf{Overall}  & \textbf{19.53} \\
\bottomrule
\end{tabular}
\caption{Percentage of false negatives requiring newly drawn boxes. Higher values indicate greater need for novel region creation rather than missed ground-truth selection.}
\label{tab:new_drawn_ratio}
\end{table}

\subsection{Cross-Dataset Variability}

\paragraph{Which datasets show strongest alignment?}
AI2D, InfographicsVQA, and MapIQ exhibit the strongest proposal alignment. AI2D achieves 99.64\% precision and 91.13\% recall (95.19 \(F_1\)), while InfographicsVQA reaches 97.72 \(F_1\). These results suggest that structured diagram layouts and clearer region boundaries support highly reliable proposal behavior.
\paragraph{Why is ChartQA challenging?}
ChartQA achieves 54.79\% precision and 42.22\% recall. Notably, 8,308 of 8,523 false negatives correspond to missed ground-truth boxes rather than newly drawn regions. This pattern suggests that evidence regions exist but are not selected by the model. Improving candidate selection strategy could significantly raise recall in chart-heavy domains.

\paragraph{Why does Circuit-VQA require more intervention?}
Circuit-VQA shows moderate precision (66.97\%) and recall (49.50\%). Over half of its false negatives (56.79\%) required newly drawn regions, suggesting incomplete candidate coverage or fine-grained component requirements in circuit diagrams.

\subsection{Human-AI Verification} 
\paragraph{Does the system primarily fail due to selection or region absence?}
The breakdown in Table~\ref{tab:fn_breakdown} shows that most false negatives (13,650/16,964) correspond to missed ground-truth boxes rather than newly drawn regions. Therefore, incomplete QA-conditioned evidence selection, rather than region absence, explains most recall gaps. Improving the evidence selection mechanism could further reduce reviewer workload.

\paragraph{How often must reviewers create entirely new boxes?}
Across all datasets, only 19.53\% of false negatives required newly drawn boxes (Table~\ref{tab:new_drawn_ratio}). In most cases, evidence regions already exist but were not proposed. This finding confirms that the review-first workflow primarily involves selecting from available regions rather than drawing from scratch.


\paragraph{Does proposal utility remain stable across heterogeneous visual domains?}
Proposal utility remains strong across heterogeneous visual domains, with performance varying by dataset. AI2D and MapIQ show high alignment, while ChartQA and Circuit-VQA exhibit lower recall. Despite these differences, the overall 80.03 \(F_1\) indicates that QA-conditioned evidence selection generalizes across charts, maps, circuits, and infographics under heterogeneous dataset formats.


\subsection{Inter-Annotator Reliability (Selection-Based Proposals)}

We assess annotation reliability for selection-based evidence proposals on 100 instances per dataset. We evaluate agreement under two complementary criteria: Complete Visual Reasoning (CVR), which requires full reasoning-chain coverage, and Core Evidence Alignment (CEA), which requires correct grounding of the primary evidence elements.

\paragraph{Do reviewers agree on proposal verification?}
Yes. Under CVR, ChartQA achieves 87.0\% agreement ($\kappa = 0.658$) and Circuit-VQA achieves 73.0\% agreement ($\kappa = 0.464$). Under CEA, agreement remains high across both datasets (83.0\% and 85.0\%, respectively). These results indicate reproducible selection-level verification across structured charts and spatially complex circuit diagrams.

We report full criterion definitions and extended discussion in Appendix~\ref{appendix:iaa}.


\begin{table}[t]
\small
\centering
\setlength{\tabcolsep}{6pt}
\begin{tabular}{lccc}
\toprule
\textbf{Dataset} & \textbf{Criterion} & \textbf{Agreement (\%)} & \textbf{$\kappa$} \\
\midrule
ChartQA        & CVR & \textbf{87.0} & \textbf{0.658} \\
ChartQA        & CEA & 83.0          & 0.589 \\
Circuit-VQA    & CVR & 73.0          & 0.464 \\
Circuit-VQA    & CEA & \textbf{85.0} & 0.540 \\
\bottomrule
\end{tabular}
\caption{Inter-annotator agreement on 100 instances per dataset under two verification criteria.}
\end{table}

\section{Demo and Reproducibility}

We release \diagrams{} as an open-source MIT-licensed repository and provide the complete review interface, including dataset loading, meta-schema normalization, QA-conditioned proposal visualization, interactive region editing, and export of reasoning-level annotations. The repository contains the unified meta-schema, dataset adapters for all evaluated datasets, and example input–output records. Researchers can load supported Diagram QA datasets, conduct human-in-the-loop verification, and export standardized attributions. Appendix~\ref{appendix:Evaluation} details the proposal utility metrics and evaluation procedure.

\section{Conclusion}

We presented \diagrams{}, a review-first framework for reasoning-level attribution in Diagram QA. The system links each question–answer pair to the complete set of visual regions required for reasoning and separates dataset-specific parsing from attribution logic through a unified meta-schema representation. By combining QA-conditioned proposal models with structured human verification, \diagrams{} shifts the workflow from manual region creation to guided review. Across six heterogeneous Diagram QA datasets, model-suggested evidence achieved 85.39\% precision and 75.30\% recall (micro-averaged) against reviewer-final selections, demonstrating strong alignment under a review-first workflow. Our analysis shows that most recall gaps arise from missed candidate selection rather than absence of regions, indicating that improved QA-conditioned selection could further reduce reviewer effort. The released framework, schema, and dataset adapters support structured reasoning-level attribution across diverse Diagram QA resources and enable grounded supervision and dataset auditing for diagram understanding.


\section*{Limitations}
\diagrams{} improves reasoning-level attribution efficiency but has several limitations. First, evidence quality depends on the underlying multimodal models, and performance may degrade for densely packed or visually cluttered diagrams that require fine-grained localization. Second, the current system represents evidence primarily as bounding boxes, which may not accurately capture thin lines, irregular shapes, or non-rectangular regions common in circuits and infographics. Third, our evaluation measures proposal utility under a review-first workflow but does not quantify absolute annotation time reduction through controlled user studies. Finally, the framework assumes structured QA inputs and does not support chain-of-thought supervision. Future work may incorporate polygon-level annotation, latency analysis, broader human studies, and structured reasoning trace integration.

\section*{Ethics Statement}

We designed \diagrams{} to support transparent and responsible human-in-the-loop annotation. All datasets used in this work originate from publicly available research benchmarks under their respective licenses. The framework does not introduce new personal or sensitive data. During annotation, we provided clear written guidelines and arbitration procedures to ensure consistency and fairness. Annotators participated voluntarily and followed standardized review protocols. Because \diagrams{} integrates multimodal models for proposal generation, users must carefully verify all model-suggested regions and QA items before finalizing annotations. The system does not assume model correctness and places decision authority entirely with the human reviewer. We release the schema specification, adapters, and evaluation scripts to promote reproducibility and transparency. We encourage users to respect original dataset licenses when redistributing derived annotations.

\section*{Acknowledgements}

We conducted this research as a collaborative effort at Arizona State University. We thank the Complex Data Analysis and Reasoning Lab within the School of Augmented Intelligence at Arizona State University for providing computational resources and institutional support. We also thank Neha Valeti and Gaurav Najpande for their assistance during the annotation phase of this project. Finally, we are grateful to the anonymous reviewers for their thoughtful feedback and constructive suggestions, which helped improve the clarity and quality of this work.

\bibstyle{acl_natbib}
\bibliography{custom} 
\clearpage

\appendix

\section{Annotation Instructions}
\label{appendix:annotation_instructions}

To ensure consistency and reliability, we provided annotators with standardized written guidelines defining task objectives, labeling rules, and arbitration procedures. Below we summarize the core instructions.

\paragraph{Objective.}
Annotators verified the correctness and clarity of each question–answer (QA) pair and identified all visual regions required to derive the answer. For each QA item, annotators selected bounding boxes corresponding to the interconnected visual elements necessary for reasoning, rather than marking only the region containing the final answer.

\paragraph{Procedure.}
\begin{enumerate}
\setlength\itemsep{0em}
\item Read the question carefully and locate all referenced visual elements in the image.
\item Use the verified answer (and answer choices, if provided) to select the bounding boxes that support the reasoning process.
\item If a question is ambiguous, inconsistent, or unanswerable from the image, flag it for arbitration rather than guessing.
\end{enumerate}

\paragraph{Annotation Policy.}
\begin{enumerate}
\setlength\itemsep{0em}
\item Two independent annotators labeled each item (six annotators participated overall).
\item A senior annotator reviewed disagreements following predefined arbitration guidelines.
\item We retained only consensus or majority-agreed annotations in the final dataset.
\end{enumerate}

\paragraph{Examples.}
We provided dataset-specific examples illustrating reasoning-level attribution across diagrams, charts, maps, circuits, and infographics.

We will release the full instruction documents, templates, and arbitration notes as supplementary materials to support transparency and reproducibility.

\section{Evaluation}
\label{appendix:Evaluation}
\subsection{Proposal Utility Metrics}

We evaluate the usefulness of model-suggested evidence regions in a review-first workflow by comparing the proposed box set \(P\) to the final reviewer-approved evidence set \(H\). 

We define:
\begin{itemize}
\setlength\itemsep{0em}
\item True positives: \(TP = |P \cap H|\) (proposed boxes retained after review),
\item False positives: \(FP = |P \setminus H|\) (proposed boxes removed by the reviewer),
\item False negatives: \(FN = |H \setminus P|\) (evidence boxes added during review).
\end{itemize}

In our logs, \(TP\) corresponds to \texttt{retained\_pred\_count}, \(FP\) to \texttt{effective\_removed\_count}, and \(FN\) to the sum of \texttt{added\_gt\_count} and \texttt{new\_drawn\_count}. 

We report micro-averaged precision \(TP/(TP+FP)\), recall \(TP/(TP+FN)\), and \(F_1\). We do not define true negatives because there is no closed set of non-evidence regions.

\subsection{Discussion} 
The results suggest that the review-first workflow is effective across multiple diagram domains but that datasets differ in whether errors arise primarily from (i) \emph{distracting proposals} (higher FP, lower precision) or (ii) \emph{coverage gaps} (higher FN, lower recall). ChartQA is the most challenging in our evaluation, with both lower precision and lower recall, consistent with the difficulty of precisely grounding evidence in dense chart elements and small text. AI2D, InfographicsVQA, and MapIQ show particularly strong performance, where proposals cover most of the final evidence with few removals. These findings motivate future improvements such as confidence-based filtering, better QA-conditioned selection strategies, and optional region refinement tools for fine-grained elements.

\section{Inter-Annotator Agreement Details}
\label{appendix:iaa}

\subsection{Evaluation Scope}

We evaluate reliability for the selection-based proposal workflow only, where the system selects evidence from pre-existing candidate regions. We exclude newly drawn regions from this analysis to isolate agreement on evidence selection quality without conflating variability introduced by bounding-box generation.

\subsection{Evaluation Criteria}

We define two complementary verification criteria:

\paragraph{Complete Visual Reasoning (CVR).}
Reviewers mark an instance as correct only when the selected regions capture the entire visual reasoning chain required to derive the answer. If any required intermediate evidence element is missing, reviewers mark the instance as incorrect.

\paragraph{Core Evidence Alignment (CEA).}
Reviewers mark an instance as correct when the selected regions capture the primary visual evidence necessary to answer the question, even if minor intermediate reasoning components are omitted. This criterion evaluates agreement on essential grounding rather than exhaustive coverage.

This design isolates agreement on evidence selection quality without introducing variability from newly drawn regions.

\section{Input and Output Data Format}
\label{appendix:data_format}

\subsection{Input Record Structure}

The tool accepts a JSON array containing one or more image records. Each record stores metadata, optional candidate regions, predicted regions, and one or more question–answer pairs. The system loads one image record at a time and saves annotations independently.

\begin{tcolorbox}[
  title=Example Input Format (Multi-Image JSON),
  colback=white,
  colframe=black,
  breakable
]
\begin{verbatim}
[
  {
    "image_uid": "img_001",
    "image_path": "path_to_image",
    "annotation_path": "path_to_json",
    "bbox": [...],
    "predicted_boxes": [],
    "questions": [
      {
        "question_text": "...",
        "answer_text": "...",
        "choices": []
      }
    ]
  },
  {
    "image_uid": "img_002",
    "image_path": "path_to_image",
    "annotation_path": "path_to_json",
    "bbox": [],
    "predicted_boxes": [],
    "questions": []
  }
]
\end{verbatim}
\end{tcolorbox}

The loader automatically detects dataset structure and converts each record into a unified internal meta-schema before rendering.

\subsection{Unified Output Meta-Schema}

After review, the system exports one JSON file per image using a standardized meta-schema. This representation ensures consistent reasoning-level attribution across heterogeneous Diagram QA datasets.

\begin{tcolorbox}[
  title=Unified Output Format (Per Image),
  colback=white,
  colframe=black,
  breakable
]
\begin{verbatim}
{
  "dataset_type": "",
  "image": "path_to_image",
  "qa": {
    "question": "...",
    "answer": "...",
    "choices": []
  },
  "annotations": [
    {
      "id": "a_1",
      "bbox": [x, y, w, h],
      "label": "",
      "meta": {
        "source": "added",
        "kind": "bbox"
      }
    }
  ],
  "metadata": {
    "annotation_path": "path_to_json",
    "ground_truth_path": "",
    "answers": {}
  }
}
\end{verbatim}
\end{tcolorbox}

This design enables multi-image ingestion while exporting standardized, per-image reasoning-level annotations.

\end{document}